\documentclass[10pt, conference, table]{IEEEtran}
\IEEEoverridecommandlockouts

\usepackage{cite}
\usepackage{amsmath,amssymb,amsfonts}
\usepackage{algorithmic}
\usepackage{graphicx}
\usepackage{textcomp}
\usepackage{hyperref}
\usepackage{enumitem}
\usepackage{booktabs}
\usepackage{graphicx}
\usepackage{arydshln} 
\usepackage{multirow}
\usepackage{pifont}
\usepackage{cleveref}
\usepackage{array}

\usepackage{xcolor}
\def\BibTeX{{\rm B\kern-.05em{\sc i\kern-.025em b}\kern-.08em
    T\kern-.1667em\lower.7ex\hbox{E}\kern-.125emX}}
    
\usepackage{fancyhdr}
\fancypagestyle{firstpage}{
  
  \fancyhf{}
  \fancyfoot[C]{\small © 2026 IEEE. Personal use of this material is permitted. Permission from IEEE must be obtained for all other uses, in any current or future media, including reprinting/republishing this material for advertising or promotional purposes, creating new collective works, for resale or redistribution to servers or lists, or reuse of any copyrighted component of this work in other works. \\ \textbf{Accepted for publication in: 2026 IEEE International Conference on Communications (ICC).}}
}
\begin{document}

\title{Quantifying  Catastrophic Forgetting in IoT Intrusion Detection Systems}
\author{\IEEEauthorblockN{Sourasekhar Banerjee\IEEEauthorrefmark{1}, David Bergqvist\IEEEauthorrefmark{1}, Salman Toor\IEEEauthorrefmark{1}\IEEEauthorrefmark{3}, Christian Rohner\IEEEauthorrefmark{1}, and Andreas Johnsson\IEEEauthorrefmark{1}\IEEEauthorrefmark{2}}
\IEEEauthorblockA{\IEEEauthorrefmark{1} Uppsala University, Department of Information Technology, Sweden\\
Email: \emph{\{sourasekhar.banerjee, salman.toor, christian.rohner, andreas.johnsson\}@it.uu.se}  
}
\IEEEauthorblockA{\IEEEauthorrefmark{2} Ericsson Research, Research Area AI, Sweden }

\IEEEauthorblockA{\IEEEauthorrefmark{3} Scaleout Systems AB, Sweden }
}

\maketitle
\thispagestyle{firstpage}
\begin{abstract}

Distribution shifts in attack patterns within RPL-based IoT networks pose a critical threat to the reliability and security of large-scale connected systems. Intrusion Detection Systems (IDS) trained on static datasets often fail to generalize to unseen threats and suffer from catastrophic forgetting when updated with new attacks. Ensuring continual adaptability of IDS is therefore essential for maintaining robust IoT network defense. In this focused study, we formulate intrusion detection as a domain continual learning problem and propose a method-agnostic IDS framework that can integrate diverse continual learning strategies. We systematically benchmark five representative approaches across multiple domain-ordering sequences using a comprehensive multi-attack dataset comprising 48 domains. Results show that continual learning mitigates catastrophic forgetting while maintaining a balance between plasticity, stability, and efficiency—a crucial aspect for resource-constrained IoT environments. Among the methods, Replay-based approaches achieve the best overall performance, while Synaptic Intelligence (SI) delivers near-zero forgetting with high training efficiency, demonstrating strong potential for stable and sustainable IDS deployment in dynamic IoT networks.

\end{abstract}


\section{Introduction}

The rapid adoption of Internet of Things (IoT) technologies has transformed the way modern societies function. It enables the pervasive connectivity across multiple devices. From smart homes and wearable sensors to critical infrastructure \cite{hassan2019current}, IoT systems form the digital backbone of emerging cyber-physical environments. However, this widespread connectivity also introduces unprecedented security challenges. Attackers can exploit the heterogeneity and resource limitations of IoT devices to compromise entire networks and disrupt essential services.
In many IoT deployments, devices operate in resource-constrained and lossy network environments. To maintain scalable and energy-efficient connectivity under such conditions, protocols like RPL (IPv6 Routing Protocol for Low-power and Lossy Networks) \cite{rfc6550}  are commonly employed. While effective for routing, their lightweight design exposes the network to a variety of attacks, for example, Blackhole (BH) \cite{kaveh2024impact}, DIS-Flooding (DF) \cite{kaveh2025factors}, Worst Parent (WP), and Local Repair (LR) attacks \cite{bergqvist2025assessing}. These attacks exploit vulnerabilities in RPL to degrade performance, increase latency, and in severe cases, partition the network. To ensure secure and reliable operation, Intrusion Detection Systems (IDS) play a critical role in identifying such malicious activities before they cause large-scale disruptions.
 \begin{figure}[!t]
    \centering
    \includegraphics[width=\linewidth]{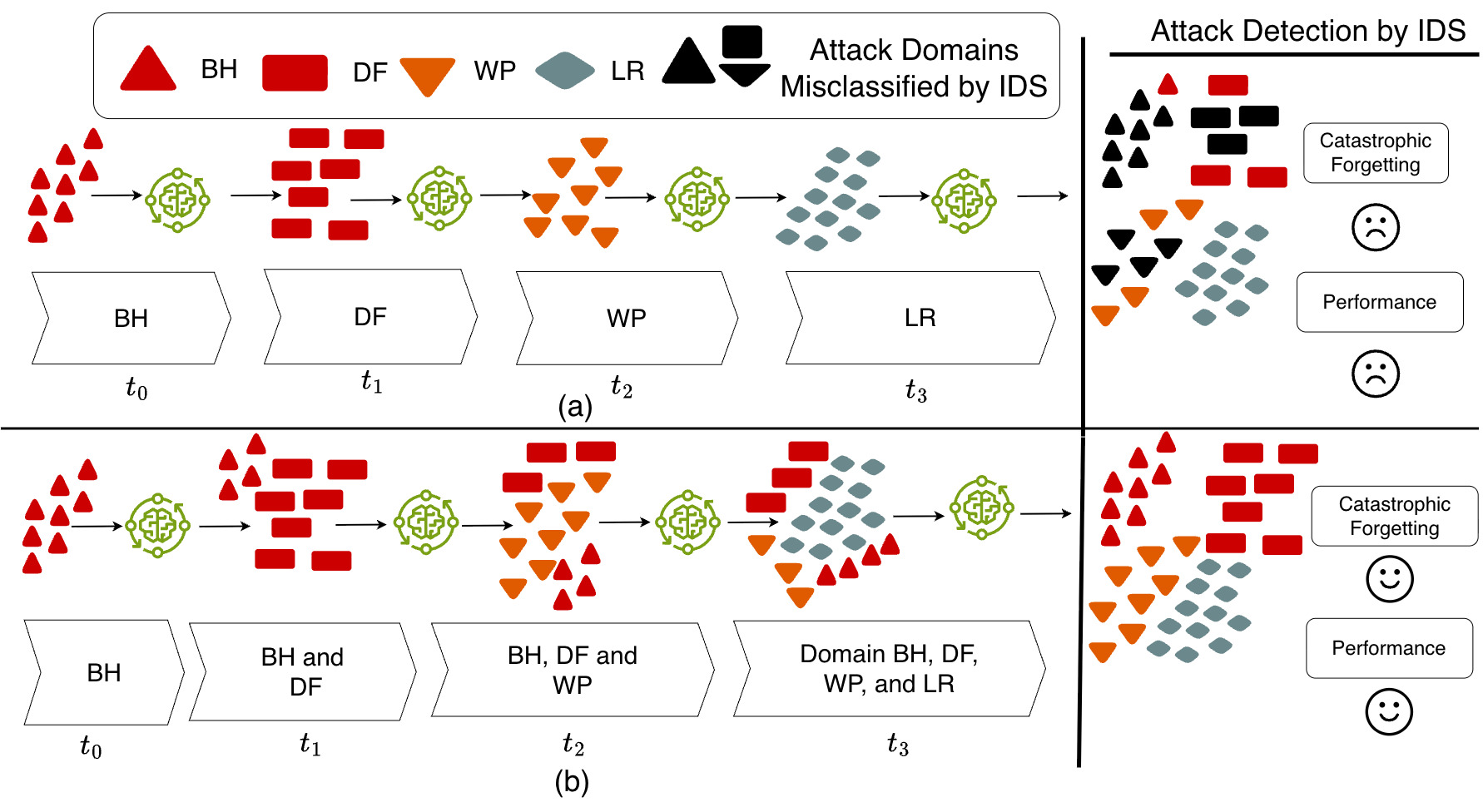}
    \caption{(a) Sequential training suffers from catastrophic forgetting, while (b) continual learning retains knowledge of prior attacks to sustain performance.}
    \label{fig: cl_wocl}
\end{figure}
Traditional IDS techniques,  such as, anomaly-based \cite{einy2021anomaly} methods, as well as Machine Learning (ML)-based IDS that rely on historical IoT traffic data to train discriminative models \cite{al2025comprehensive}, all face generalization challenges when exposed to unseen attacks \cite{kaveh2023impact}. Since these models are typically trained on a static scenario, they become vulnerable to novel attacks that exhibit patterns significantly different from the training data \cite{kaveh2025factors, kaveh2023impact}. Updating an ML-based IDS model with new attack data introduces the stability-plasticity dilemma. It means, the IDS model must remain plastic enough to learn new cyber attacks while being stable enough to retain knowledge of previously learned ones. Failure to balance this trade-off often results in catastrophic forgetting \cite{prasath2022analysis}, where the IDS adapts to new attacks but loses detection capability on earlier ones, which limits its effectiveness in dynamic IoT environments.  As shown in \Cref{fig: cl_wocl}(a), sequential training on different attack types (e.g., BH, DF, WP, LR at timestamp $t_0$ to $t_3$ results in increased misclassification of earlier attacks, revealing the impact of catastrophic forgetting which leads to poor performance.

To overcome these limitations, we propose a domain Continual Learning (CL)--based IDS for RPL-based IoT networks that mitigates the forgetting problem by retaining knowledge of previously learned attacks across successive training rounds. For example, as shown in \Cref{fig: cl_wocl}(b), the CL-based IDS mitigates catastrophic forgetting by retaining knowledge from previous attack domains through sample replay or parameter preservation. Consequently, after training on all domains, the IDS achieves improved performance across all attacks.  In our approach, the IDS model, based upon features derived from RPL statistics collected in the network, is incrementally updated as new types of attacks emerge, without retraining from scratch, but  retaining knowledge of previously learned attacks. This is particularly important for IoT, where the threat landscape evolves continuously and static IDS models suffer from catastrophic forgetting when fine-tuned on new data. A key reasons behind this forgetting is the distributional shift across different attack types, despite the network status is normal (0) or under attack (1). For example, BH, DF, WP, and LR attacks manipulate different protocol mechanisms, leading to heterogeneous traffic patterns even though they all share the same attack label. Conventional binary classifiers \cite{kaveh2025factors, kaveh2023impact} trained on static datasets struggle to maintain high performance across such evolving distributions. By framing the problem as domain continual learning, we treat each attack type or variant as a separate domain and train the IDS sequentially. This enables the model to adapt to new distributions while retaining previously acquired knowledge. Unlike conventional approaches, CL explicitly balances \emph{plasticity} and \emph{stability}. Moreover, we introduce \emph{training efficiency} as a third dimension in evaluating IDS performance, capturing the model’s ability to learn effectively without excessive computational cost. Together, these three factors form a critical trade-off in designing robust IDS solutions for dynamic IoT environments.

\noindent\textbf{\textit{Contributions.}} In this paper we make the following contributions, \textbf{\textit{(i)}} We formulate intrusion detection for RPL-based IoT networks as a domain continual learning problem, and develop a method-agnostic IDS framework that can be integrated with different continual learning methods.
\textbf{\textit{(ii)}} We systematically benchmark 5 representative continual learning methods, namely Elastic Weight Consolidation (EWC), Synaptic Intelligence (SI), Learning without Forgetting (LwF), Experience Replay (Replay), and Generative Replay (GR) across multiple domain-ordering sequences, to analyze their impact on \emph{stability}, \emph{plasticity}, and \emph{training efficiency} relative to a non-continual baseline and provide guidance on which is more suitable for RPL-based IDS.
\textbf{\textit{(iii)}} We construct a comprehensive multi-attack dataset by extending \cite{kaveh2025factors} with two additional RPL attacks (Worst Parent and Local Repair), multiple behavioral variants, and network sizes; yielding 48 distinct domains, and publicly release the dataset and code \cite{uu-core_IoT_Attacks_IDS} to support reproducibility.

\section{Related Work} 
We group the related works into two categories such as, conventional IDS and continual IDS, discussed below.

Conventional IDS approaches, whether rule-based \cite{einy2021anomaly} or data-driven \cite{al2025comprehensive,kaveh2025factors}, are trained offline on static datasets. Signature-based systems fail to detect zero-day attacks \cite{prasath2022analysis}, anomaly-based systems suffer from high false alarms and poor scalability \cite{mchugh2001intrusion}, and ML models, while effective at capturing sequential traffic patterns with LSTM/BiLSTM, generalize poorly to unseen or evolving threats \cite{kaveh2023impact}. Together, these static IDS are not suitable for dynamic IoT environments where attackers continuously adapt their strategies.

To overcome this limitation, recent research has explored CL for IDS. \cite{kim2025intrusion} proposed domain-incremental IDS with episodic memory. \cite{zhang2025continual} addressed concept drift via selective forgetting. While these methods improve adaptability, they often overlook the stability–plasticity–efficiency trade-off crucial for deployment in resource-constrained IoT networks.

From the literature we identified that static approaches are unable to  evolving attacks, whereas while CL-based approaches enhance adaptability they  lack focus on IoT-specific stability-plasticity-efficiency constraints. This motivates our work and we formulate IDS as a continual learning problem, explicitly balancing plasticity, stability, and efficiency, and evaluate it across multiple evolving IoT attack domains.

\section{Background }  \label{sec: background}
In this section, we present the overview of the network model and describe the IoT  attacks and their variants used in our experiments.

\noindent \textbf{Network Model: } As in our previous work \cite{kaveh2025factors},  we consider an RPL-based network, where RPL is a IPv6 routing protocol for Low-Power and Lossy Network \cite{rfc6550}. RPL organizes IoT devices into a Destination-Oriented Directed Acyclic Graph (DODAG) rooted at a sink node (yellow round node in \Cref{fig: end_to_end}). The sink node typically has higher computation and storage capabilities and serves as the primary observation point for monitoring and intrusion detection. IoT nodes exchange RPL control messages such as, DODAG Information Solicitation (DIS), DODAG Information Object (DIO), and Destination Advertisement Object (DAO) to establish and maintain forwarding across the network. As in previous work, we let the sink aggregate statistics from all nodes, to be used for intrusion detection. 
\noindent\textbf{Attacks and Variants: }  We considered 4 RPL-based IoT network attacks, each having 3 different variants. 

\noindent\textbf{\textit{1. Blackhole (BH) \cite{kaveh2023impact}:}} A malicious node advertises an artificially low rank to attract traffic, which is then dropped, disrupting packet forwarding.

\noindent\textbf{\textit{2. DIS-Flooding (DF)\cite{kaveh2025factors}:}} an attacker repeatedly broadcasts DIS messages, forcing nodes to respond with DIOs and overwhelming the network with control traffic.

\noindent\textbf{\textit{3. Worst Parent (WP) \cite{bergqvist2025assessing}:}} a compromised node manipulates rank values, causing neighbors to select suboptimal routes and degrading performance.

\noindent\textbf{\textit{4. Local Repair (LR)  \cite{bergqvist2025assessing}:}} an attacker repeatedly triggers false local repairs, increasing routing overhead and delay.

Each of these attacks can manifest in 3 behavioral variants \cite{kaveh2025factors}: base, on–off, and gradual change. In the base variant, the attack starts suddenly and continues without changing. In the on–off variant, the attacker switches between normal and malicious behavior according to a predefined pattern. Finally,  gradual change is similar to the base variant, but instead of an abrupt start, is gradually changing the attack intensity.

\section{Problem Statement and Formulation} \label{sec: problem_formulation}

IoT networks are inherently dynamic, with evolving routing patterns and attack strategies that cause distributional shifts in the observed network data, the basis for an ML-based IDS. Such shifts may occur gradually (e.g. due to configuration changes) or abruptly (e.g. when new attack variants emerge), leading to changes in the underlying data distribution over time. This phenomenon is also referred to as concept drift. Static IDS models trained offline fail to generalize to these evolving conditions, while naive fine-tuning on new data causes catastrophic forgetting of earlier attack patterns. Consequently, IDS performance degrades over time under concept drift. To address this challenge, we formulate IDS training as CL problem, where the goal is to gradually learn from evolving attacks while retaining previously acquired knowledge.

\noindent\textbf{\textit{Formulation:}}
The learning process is structured into \( T \) sequentially arriving \emph{domains}, denoted by \( \{ D_t \}_{t=1}^T \). Each domain corresponds to a distinct attack scenario or variant (i.e. Blackhole, DIS-Flooding, Worst Parent, Local Repair, and their behavioral variants or variation in network size), and is observed only once in a streaming manner. Formally, each domain \( D_t \) consists of a dataset 
\(D_t = \left\{ (x_i^t, y_i^t) \right\}_{i=1}^{N_t},\)
where \( x_i^t \sim P_t(X) \) represents the input feature vector \cite{kaveh2025factors} sampled from a domain-specific distribution \( P_t(X) \), where X refers to the input feature vector representing characteristics of network traffic. And \( y_i^t \in Y = \{0,1\} \) is the corresponding label indicating normal (\(0\)) or attack (\(1\)) traffic.   While the label space remains fixed across domains, the feature distributions differ (\( P_t(X) \neq P_{t'}(X) \) for \( t \neq t' \)) which reflects domain shifts over time.

Let \( f_\theta : X \to [0,1] \) be a  IDS model with parameters \( \theta \). At each step, the IDS learner updates its parameters using the current domain data \( D_t \) and some auxiliary information \( \mathcal{A}_{t-1} \) retained from previous domains:
\((\theta_{t-1}, D_t, \mathcal{A}_{t-1}) \; \mapsto \; (\theta_t, \mathcal{A}_t).\) The auxiliary information \( \mathcal{A}_t \)  depends on the chosen CL method. For example, for \textbf{\textit{regularization-based methods}} $\mathcal{A}_t$ is the importance weights or Fisher information, whereas for \textbf{\textit{knowledge distillation methods}}, it is the teacher outputs or model snapshots. Finally, for \textbf{\textit{replay-based methods}}, it is samples from past domain or generated samples. The overall objective is to minimize the cumulative expected loss over all domains encountered so far, while respecting constraints on memory usage and forgetting:
\begin{equation}
\min_{\{\theta_t, \mathcal{A}_t\}_{t=1}^T} \; \frac{1}{T} \sum_{t=1}^{T} \mathcal{R}_t(\theta_T),
\quad \text{s.t.} \quad
|\mathcal{A}_t| \le B, \quad \overline{S}_t \le \epsilon ,
\label{eq: main_opt}
\end{equation}
where
\( \mathcal{R}_t(\theta) = \mathbb{E}_{(x,y)\sim D_t}\bigl[ \ell(f_\theta(x), y) \bigr]
\) is the expected loss on domain \( D_t \), \( B \) is the memory budget, and \( \overline{S}_t \) is average forgetting on previous domains. $\epsilon$ is a forgetting threshold.

\section{System Model} \label{sec: system_model}

\begin{figure}[!htbp]
    \centering
    \includegraphics[width=\linewidth]{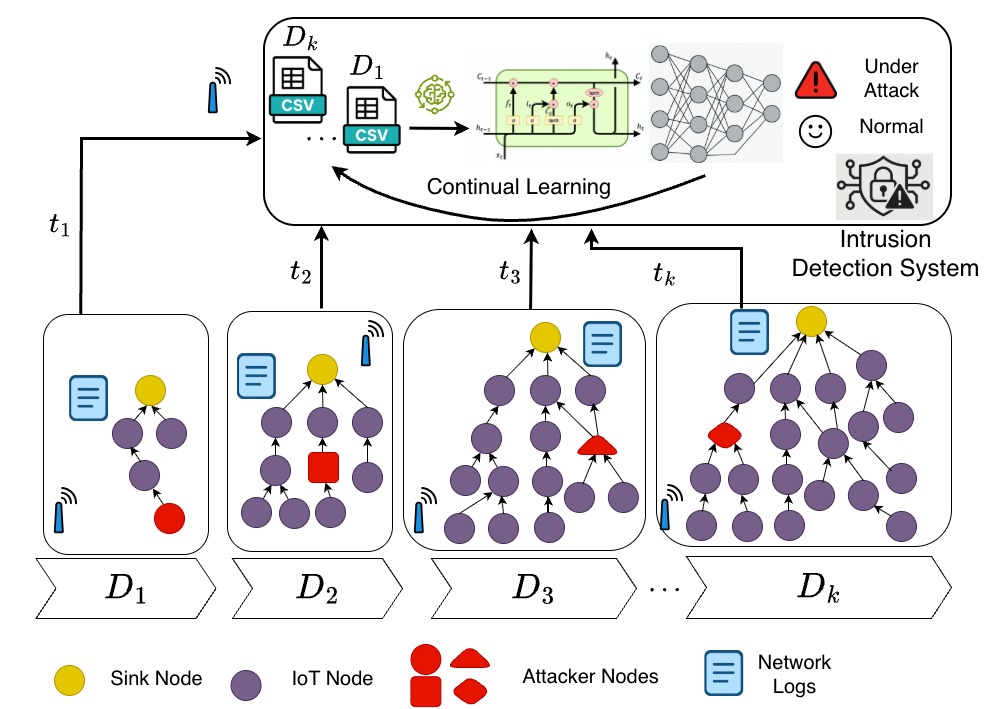}
    \caption{System model of continual learning based intrusion detection system. }
    \label{fig: end_to_end}
    \vspace{-0.1in}
\end{figure}

 We propose a robust IDS for RPL-based IoT networks, utilizing continual learning to enhance detection capability of network attacks adaptively over time. The system architecture and sequences of domains are illustrated in \cref{fig: end_to_end}.

\subsection{Network, domains, and data collection}

\begin{figure}[!t]
    \centering
    \scriptsize
    \includegraphics[width=0.9\linewidth]{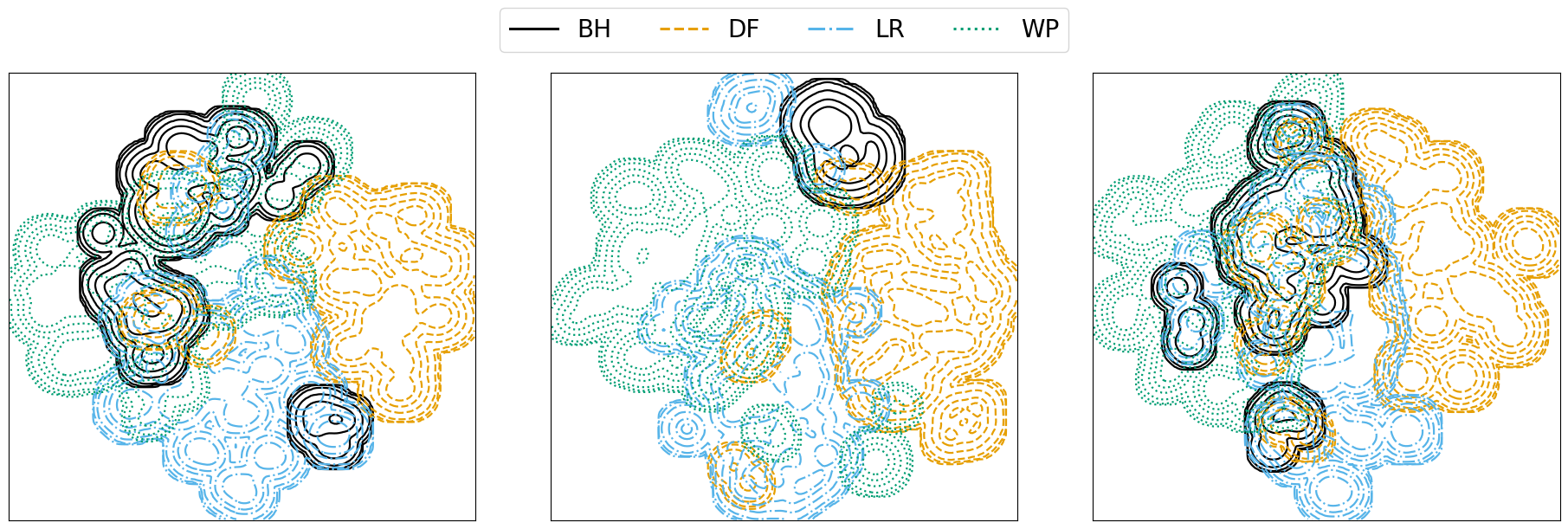}
    \caption{T-SNE 2D-visualizations of attack domains for four IoT network attacks—Blackhole (BH), DIS-Flooding (DF), Local Repair (LR), and Worst Parent (WP)—under three attack types: base (left), on-off (center), and gradual decrease (right). Network comprises 20 nodes, simulated using Cooja \cite{bergqvist2025assessing}.}
    \label{fig: distribution_shift}
\end{figure}

The IoT network in this work (see \cref{fig: end_to_end}) consists of three types of nodes such as, IoT nodes (purple) that generate data, a Sink node (yellow) that collects logs, and Attacker nodes (red) that disrupt normal operation by  modifying control messages. The network can vary in size, for example 5, 10, 15, or 20 nodes, and each configuration is treated as a separate domain, which may be similar or disjoint depending on the scenario.
\cref{fig: distribution_shift} provides an illustration using t-SNE plots of four RPL attack domains (BH, DF, LR, WP) under base, on–off, and gradual decrease. DF remains the most distinct, with minimal overlap with other attacks, whereas BH, LR, and WP share little overlapping regions that reflects these attacks has some similarity on their distributions. This clearly demonstrates the challenge of distributional shifts.  

At discrete time intervals (see \cref{fig: end_to_end}) \( t_1, t_2, \ldots, t_k \) network logs are collected at the sink node for each domain. These logs include control message exchanges (DIS, DIO, DAO), routing updates (rank), and traffic statistics (e.g., total RPL messages sent), capturing both normal and under attack. The logs are then transferred to a server and converted into CSV files, forming domain-specific structured datasets used for subsequent analysis. A detailed description of the collected statistics is provided in the experimental setup (see \cref{sec: experimental_setup}).

\subsection{Continual Learning integrated into the IDS} 

The system adopts a domain incremental learning setup, where datasets from domains $D_1$ through $D_k$, are observed sequentially over time, each corresponding to a specific network configuration or attack variants. Instead of retraining from scratch, the IDS incrementally updates its model as new domains arrive, enabling it to integrate knowledge of emerging attacks while preserving previously acquired information.
To achieve this, we employ a method-agnostic continual learning framework that supports multiple strategies. Specifically, we use regularization-based methods (e.g., Elastic Weight Consolidation (EWC) \cite{kirkpatrick2017overcoming}, Synaptic Intelligence (SI) \cite{zenke2017continual}) to protect important parameters, distillation-based methods (e.g., Learning without Forgetting (LwF) \cite{li2017learning}) to transfer knowledge without revisiting old data, and replay-based approaches \cite{robins1995catastrophic, shin2017continual} that store a small set of representative samples from past domains to stabilize learning. This framework can be applied to any sequential model architecture, making the IDS adaptable to evolving threats in a flexible and scalable manner.
\section{Experimental Setup} \label{sec: experimental_setup}

\noindent\textit{\textbf{Datasets:}} We build upon the attack definitions and variants introduced in \cref{sec: background} to construct a comprehensive dataset for evaluating IDS performance under evolving IoT attacks. Specifically, we simulate 4 RPL-based attack types (Blackhole, DIS-Flooding, Worst Parent, and Local Repair), each manifested in three variants (Base, On–Off, and Gradual Decrease) and deployed across 4 different network sizes (5, 10, 15, and 20 nodes). This systematic combination yields 48 distinct domains, covering both small-scale and large-scale IoT deployments. Each domain comprises 20 simulation runs, ensuring diversity and robustness in the generated data. The details of the attacks is available in \cite{bergqvist2025assessing}. The complete dataset, along with preprocessing scripts and IDS training pipelines, is publicly available at \cite{uu-core_IoT_Attacks_IDS}.

\noindent\textbf{\textit{Feature Extraction from RPL Control Messages: }}
To support a data-driven IDS at the sink, we configure the network so that RPL control messages are visible at the sink. Following prior works \cite{finne2021multi, kaveh2023impact, kaveh2025factors}, each IoT node sends a UDP packet every minute with 7 values: (1) current rank, (2) DIS sent, (3) DIO sent, (4) DAO sent, (5) DIS received, (6) DIO received, and (7) total RPL messages sent. The sink logs each packet with timestamp and sender. For network-wide behavior, the sink computes the mean ($\mu$) and standard deviation ($\sigma$) of these 7 attributes per minute. This yields a 14-dimensional feature vector,
\(x_t = [\mu_1, \ldots, \mu_7, \sigma_1, \ldots, \sigma_7],
\) which is min--max normalized and used to train and evaluate sequential ML models.  

\noindent\textbf{\textit{Model: }}
We selected an LSTM architecture as the base model due to its proven efficacy in capturing sequential traffic patterns in RPL-based networks \cite{kaveh2025factors}. Our model has an LSTM layer to capture temporal dependencies from input sequences, followed by a two-layer feedforward head for prediction. The LSTM encodes the input into hidden representations, from which the last time step is taken as a compact feature. This feature passes through a fully connected layer with ReLU activation, dropout for regularization, and a final linear layer that outputs logits corresponding to the target classes normal (0) or attack (1).

\noindent\textbf{\textit{CL Methods and Baselines: }}
We compared a naive fine-tuning baseline (W/O CL) against five CL methods covering all major categories: Regularization (EWC \cite{kirkpatrick2017overcoming}, SI \cite{zenke2017continual}), Distillation (LwF \cite{li2017learning}), and Rehearsal (Experience Replay \cite{robins1995catastrophic}, Generative Replay \cite{shin2017continual}). Experiments were conducted with a unified training setup such as, 0.001 learning rate, 100 epochs, and a batch size of 256 to ensure a fair comparison across domain sequences. Further implementation details are available in \cite{uu-core_IoT_Attacks_IDS}

\noindent\textbf{\textit{CL Scenarios: }}To examine how domain exposure sequences influence continual learning performance, we consider four ordering scenarios: Best-to-Worst (B2W), Worst-to-Best (W2B), Random, and Adversarial (Toggle). These simulate diverse learning conditions an IDS may face. B2W starts with highly generalizable domains and moves to harder ones, representing an optimistic scenario where early learning builds a strong foundation. W2B begins with the least generalizable domains, creating a pessimistic scenario that may cause poor initial representations and greater forgetting. Random presents domains in arbitrary order, reflecting real-world unpredictability. Adversarial toggles between highly and poorly generalizable domains to stress-test the model’s stability, adaptability, and robustness to abrupt distributional shifts. Domain generalizability is assessed based on how well a model trained on that domain performs on other unseen domains. High generalizability indicates transferable patterns, while low suggests limited cross-domain utility.

\noindent\textbf{\textit{Evaluation Metrics: }}
We assess the impact of catastrophic forgetting on IDS using four evaluation metrics--- performance (F1-Score/AUC)\cite{kaveh2025factors}, plasticity, stability (BWT), and training-efficiency.

\noindent\textit{1) Performance:}
Average performance after $T$ domains:
\begin{align} \label{eq:performance}
    \overline{Perf_T}=\frac{1}{T}\sum_{t=1}^{T}Perf_{T,t}, \in [0,1]
\end{align}
where $Perf_{T,t}$ is performance on domain $t$ after learning all $T$ domains.

\noindent\textit{2) Plasticity:} It measures the model’s ability to learn new domains as they arrive. The mean plasticity is defined as
\begin{align}
\bar P = \frac{1}{T-1}\sum_{j=2}^{T}\big(Perf_{j,j}-Perf_{j-1,j}\big)   
\end{align}
where \(Perf_{i,j}\) denotes the performance on domain \(j\) after training up to domain \(i\). 

\noindent\textit{3) Stability:} measures the model’s ability to retain knowledge of previously learned domains as training progresses. We quantify stability using the Backward Weight Transfer (BWT) metric \cite{lopez2017gradient}, defined as,
\begin{align}
    \bar S = \frac{1}{T-1}\sum_{t=2}^{T}  \underbrace{\frac{1}{t-1}\sum_{i=1}^{t-1}\big(Perf_{t,i}-Perf_{i,i}\big)}_{BWT_t}
\end{align}
where \(Perf_{t,i}\) denotes the performance on domain \(i\) after training up to domain \(t\), and \(Perf_{i,i}\) is the performance on domain \(i\) immediately after learning it. The inner term $BWT_t$ captures the retention of previously acquired knowledge after training on domain t, while the outer average aggregates this effect across all domains to yield the mean stability $\bar S$
\begin{equation} \label{eq: TE}
e_t(a) = \frac{\mathrm{TT}_{\min}(t)}{\mathrm{TT}_a(t)}, 
\text{TE}(a) = \left( \prod_{t=1}^{T} e_t(a) \right)^{1/T}\in(0,1]
\end{equation}
\noindent\textit{4) Training-efficiency:}
In \cref{eq: TE}, let $\mathrm{TT}_a(t)$ denote the training time of method $a$ on domain $t$, and $\mathrm{TT}_{\min}(t)$ the best (minimum) training time achieved by any method on the same domain. The per-domain efficiency ratio is $e_t(a)$ which is aggregated across domains using the geometric mean $E(a)$.
In all the metrics, higher values indicate better performance, as denoted by ($\uparrow$) in the results given in \cref{tab:average_combined}, and \ref{tab: bwt_te}

\section{Results and Discussions}
\subsection{Performance of the model} In \cref{tab:average_combined}, we report the average F1/AUC (based on \cref{eq:performance}) of the different CL methods after training on 48 domains, aggregated over four orderings (Random, B2W, W2B, Toggle) and three attack variants: base, on-off, and gradual decrease. From the results we observed, Replay is the strongest method across all attacks, confirming that keeping and reusing past samples works best when domains change. EWC and SI provide steady but moderate gains over the W/O CL baseline, yet remain behind Replay, which suggests weight regularization alone is not sufficient. LwF and Generative Replay offer only small improvements and sometimes comparable to baseline.

By attacks, DIS-Flooding is easy to detect (high AUC) with Replay or Regularization based methods (EWC and SI). For LR attacks detection, our CL based IDS is benefited mostly from Replay method. WP and BH are the most challenging attacks to be detected by the IDS. However, Replay-based method gives \(\approx 70\% \) AUC for worst parent attacks and Blackhole attack detection. 

\begin{table}[!htbp]
    \centering
    \caption{Average performance across four scenarios (Random, B2W, W2B, Toggle) for each method and attack type.}
    \label{tab:average_combined}
    \resizebox{\linewidth}{!}{
    \begin{tabular}{p{1.3cm}|cc|cc|cc|cc}
        \toprule
        \multirow{2}{*}{\textbf{Method}} & \multicolumn{2}{c}{\textbf{Blackhole (BH)}} & \multicolumn{2}{c}{\textbf{DIS-Flooding (DF)}} & \multicolumn{2}{c}{\textbf{Local Repair (LR)}} & \multicolumn{2}{c}{\textbf{Worst Parent (WP)}} \\
        \cmidrule(lr){2-3} \cmidrule(lr){4-5} \cmidrule(lr){6-7} \cmidrule(lr){8-9}
        & \textbf{$F1 \uparrow$} & \textbf{$AUC \uparrow$} & \textbf{$F1 \uparrow$} & \textbf{$AUC \uparrow$} & \textbf{$F1 \uparrow$} & \textbf{$AUC \uparrow$} & \textbf{$F1 \uparrow$} & \textbf{$AUC \uparrow$} \\
        \midrule
        \textbf{W/O CL}               & 0.30 & 0.49 & 0.78 & 0.96  & 0.63 & 0.75 & 0.45 & 0.57 \\
        \textbf{Replay}              & \cellcolor{yellow}0.64 &  \cellcolor{yellow}0.69     & \cellcolor{yellow}0.97 & \cellcolor{yellow}1.00   & \cellcolor{yellow}0.88 & \cellcolor{yellow}0.95   & \cellcolor{yellow}0.63 & \cellcolor{yellow}0.70   \\
  
        \textbf{EWC}                 & 0.52 & 0.64   & 0.82 & \cellcolor{yellow}1.00 & 0.63 & 0.74 & 0.48 & 0.58 \\
        
        \textbf{SI}                  & 0.42 & 0.61   & 0.88 & \cellcolor{yellow}1.00  & 0.58 & 0.77 & 0.47 & 0.59 \\
        
        \textbf{LwF}                 & 0.33 & 0.52   & 0.71 & 0.81  & 0.54 & 0.72 & 0.43 & 0.58 \\

        \textbf{GR}  & 0.30 & 0.50 & 0.77 & 0.94    & 0.59 & 0.77   & 0.42 & 0.58   \\
        \bottomrule
    \end{tabular}
    }
\end{table}

\subsection{Catastrophic Forgetting}
\begin{figure}[!htbp]
    \centering
    \scriptsize
    \includegraphics[width=\linewidth]{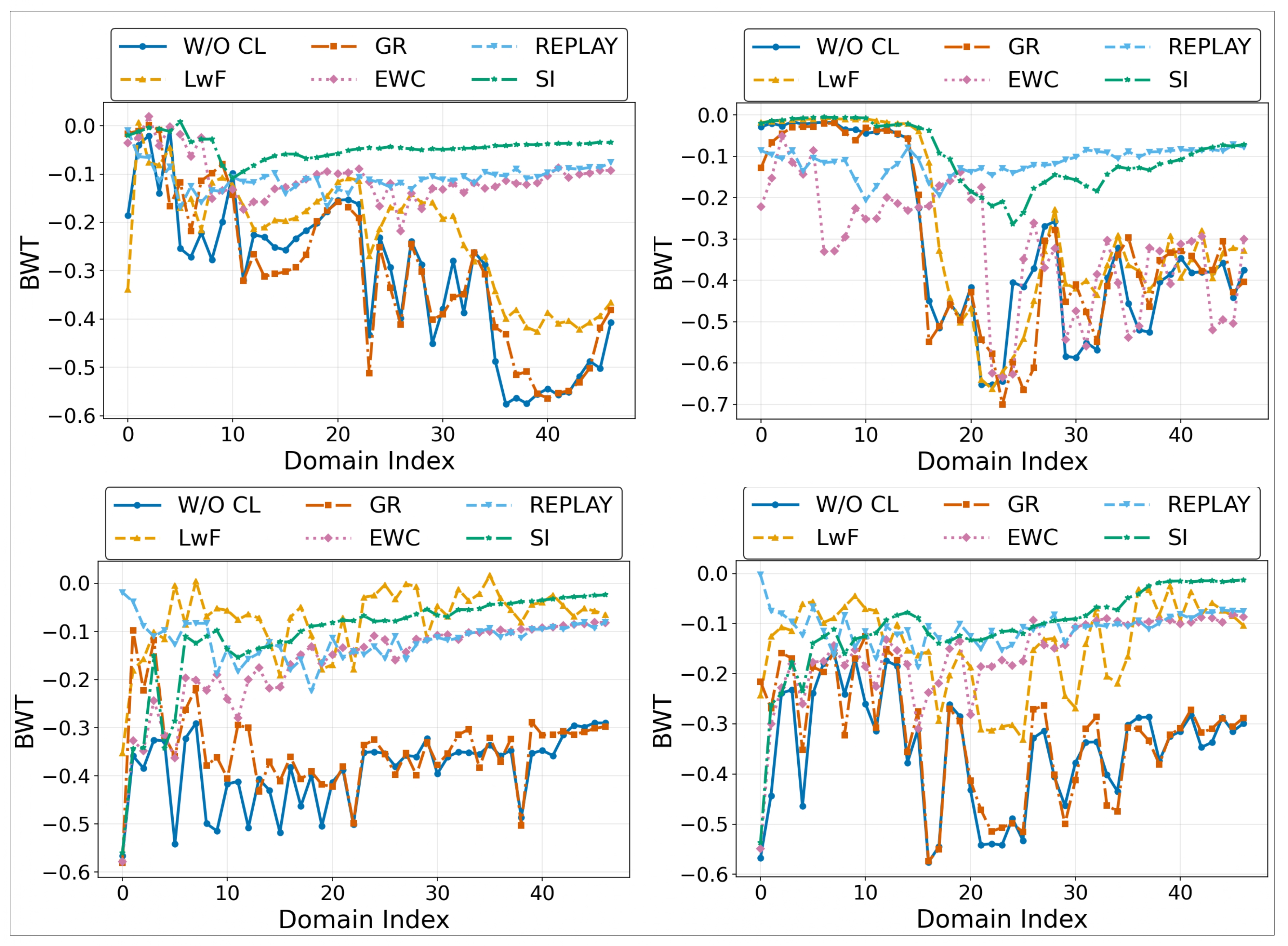}
    \caption{Change of BWT per domain for domain sequence Random (Top left), B2W (Top right), W2B (Bottom left), Toggle (Bottom right).}
    \label{fig: bwt_rate}
\end{figure}

In \cref{fig: bwt_rate}, we evaluate the extent of catastrophic forgetting for each CL method using the BWT metric. A negative BWT 
indicates that the model has experienced catastrophic forgetting, whereas  BWT $\ge 0$ indicates that the model avoids forgetting. Across the four scenarios in \cref{fig: bwt_rate} we observe a consistent pattern of chatastropic forgetting across domains and scenario; Random (top left), B2W (top right), W2B (bottom left), and Toggle (bottom right). This reveals the robustness (or vulnerability) of each method under different domain orderings. The baseline, W/O CL, exhibits the strongest negative BWT and forgets rapidly. EWC keeps BWT values much closer to zero across most sequences, indicating that regularization preserves earlier knowledge. Replay and SI stay near zero throughout, which reflects good retention with steady adaptation. GR and LwF improve over the baseline but remain variable and less stable across all domain sequences. Overall, Replay and SI provide the most reliable retention, EWC offers moderate protection at the cost of adaptation, and GR and LwF do not fully stabilize forgetting across sequences.

\begin{table}[!htbp]
\centering
\caption{Average Stability ($\bar{S}$), Plasticity ($\bar{P}$), and Training Efficiency (TE) after training on $48$ domains.}
\label{tab: bwt_te}
\resizebox{\linewidth}{!}{
\begin{tabular}{p{0.6cm} ccc|ccc|ccc|ccc}
\toprule
\multirow{2}{*}{\textbf{Method}} & \multicolumn{3}{c}{\textbf{Random}} & \multicolumn{3}{c}{\textbf{B2W}} & \multicolumn{3}{c}{\textbf{W2B}} & \multicolumn{3}{c}{\textbf{Toggle}} \\
\cmidrule(lr){2-4} \cmidrule(lr){5-7} \cmidrule(lr){8-10} \cmidrule(lr){11-13}
& $\bar{S} \uparrow $ & $\bar{P} \uparrow $ & $TE \uparrow$ & $\bar{S} \uparrow $ & $\bar{P} \uparrow $ & $TE \uparrow $& $\bar{S} \uparrow $ & $\bar{P} \uparrow $ & $TE \uparrow$ & $\bar{S} \uparrow $ & $\bar{P} \uparrow $ & $TE \uparrow$ \\
\midrule
\textbf{W/O CL}              & -0.32 & 0.25 & \cellcolor{yellow}0.94 & -0.31 & 0.13 & \cellcolor{yellow}0.96 & -0.38 & \cellcolor{yellow}0.40 & 0.85 & -0.35 & \cellcolor{yellow}0.31 & \cellcolor{yellow}0.96 \\
\textbf{Replay}              & -0.11 & \cellcolor{yellow}0.31 & 0.54 & -0.11 & \cellcolor{yellow}0.32 & 0.52 & -0.11 & 0.31 & 0.66 & -0.11 & \cellcolor{yellow}0.31 & 0.57 \\
\textbf{EWC}                 & -0.15 & 0.08 & 0.18 & -0.28 & 0.19 & 0.22 & -0.17 & 0.16 & 0.20 & -0.17 & 0.14 & 0.18 \\
\textbf{SI}                  & \cellcolor{yellow}-0.06 & 0.03 & 0.83 & \cellcolor{yellow}-0.08 & 0.03 & 0.80 & \cellcolor{yellow}-0.09 & 0.06 & \cellcolor{yellow}0.92 & \cellcolor{yellow}-0.09 & 0.05 & 0.78 \\
\textbf{LwF}                 & -0.22 & 0.07 & 0.64 & -0.26 & 0.08 & 0.61 & -0.10 & 0.14 & 0.76 & -0.16 & 0.13 & 0.64 \\
\textbf{GR}  & -0.28 & 0.24 & 0.42 & -0.29 & 0.13 & 0.40 & -0.35 & 0.38 & 0.51 & -0.32 & 0.29 & 0.42 \\
\bottomrule
\end{tabular}
}
\end{table}

\subsection{Stability, Plasticity, and Efficiency Trade--off}  
In \cref{tab: bwt_te}, we reported mean Stability (\(\bar{S}\)), Plasticity (\(\bar{P}\)), and Training Efficiency (TE) over 48 domains across four orderings (Random, B2W, W2B, Toggle). We observe Replay consistently achieves a favorable balance across all ordering by maintaining relatively high stability (\(\bar{S} = -0.11\)), strong plasticity (\(\bar{P} \approx 0.31\)), and moderate efficiency (0.52 to 0.66). Though Replay has higher memory and computational overhead but leveraging real samples, Replay effectively revisits past distributions and consolidates prior knowledge. Therefore it increase adaptability to new domains. On the other hand, GR improves plasticity under W2B (\(\bar{P} = 0.38\)) but suffers from poor stability (\(\bar{S} = -0.35\)) and lower efficiency, because generating synthetic traffic samples incurs additional cost while failing to capture the high-dimensional, temporal structure of real IoT traffic. Eventually GR is adding noise which reflects in poor stability and plasticity of the model.   

Among regularization-based methods, EWC and SI enhance stability by constraining updates to important parameters, but this stability gain comes at the expense of plasticity and efficiency. EWC shows moderate stability (\(\bar{S} \in [-0.15, - 0.28]\)) and low plasticity (\(\bar{P} \in [0.08, 0.19]\)) with the lowest efficiency (0.18 to 0.22) due to repeated Fisher matrix estimation, while SI achieves the best stability among regularization methods (\(\bar{S} = -0.06\)) and remains computationally efficient because of its online importance accumulation, though its plasticity is extremely low (\(\bar{P} \leq 0.06\)). LwF gives neither strong stability nor plasticity, with efficiency moderately affected by distillation overhead and performance remaining close to the baseline. The baseline (W/O CL) is the most training efficient ($TE$ $\in$ [0.85, 0.96]) due to its simple sequential fine-tuning, but exhibits severe forgetting (\(\bar{S}\)$\in$[-0.31, -0.38]).  

From an IoT security practitioner’s perspective, considering all scenarios (\cref{tab:average_combined}) and different attack types (\cref{tab: bwt_te}),  Replay is the preferred CL strategy, requiring data storage from previous domains, as it consistently achieves superior performance and mitigates forgetting in evolving attacks. Under stringent storage constraints typical in IoT scenarios, SI offers a practical alternative. However, SI’s strong emphasis on stability at minimal memory cost limits plasticity, resulting in slower adaptation to new or evolving attack patterns. This occurs because SI constrains parameter updates to preserve previously learned knowledge, reducing the model’s ability to reconfigure its internal representations when faced with novel threats. Consequently, while SI effectively mitigates catastrophic forgetting, it introduces a trade-off between stability and plasticity, making it less suitable for environments where attack dynamics change rapidly.

\section{Conclusions and Future Scope}

In this work, we formulated RPL-based IoT intrusion detection as a domain continual learning problem and benchmarked five representative CL methods across 48 domains encompassing multiple attack types, variants, and orderings. Our analysis confirms the fundamental trade-off between plasticity, stability  and training efficiency--a non-negotiable requirement for any solution intended for real-world IoT networks. The results demonstrate that Replay-based methods, while offering the best balance of stability and plasticity, do so at a significant cost to training efficiency. CL-methods like Synaptic Intelligence, however, emerged as a well-rounded alternative, balancing all three criteria effectively. For further benchmarking, we release our IoT network attack dataset for public use.This work relies on simulation environments, which lack real-world physical noise. Moreover, while we analyzed algorithmic efficiency, we did not verify energy or memory consumption on physical hardware. Future work will focus on designing an autonomous, lightweight replay strategy , validated through real-time deployment and comprehensive resource profiling on physical IoT testbeds.

\section*{Acknowledgment}
This research has been supported by the Swedish Governmental Agency for Innovation Systems (VINNOVA) through the project Robust IoT Security: Intrusion Detection Leveraging Contributions from Multiple Systems (2023-02982), as well as the Swedish Civil Contingencies Agency (MSB) through the Robust IoT project (2018-12526). We also express our gratitude to Dr. José Mairton Barros da Silva Júnior for his valuable comments.

\bibliographystyle{IEEEtran}
\bibliography{bibtex/bib/bibliography}

\end{document}